# SKIN LESION DETECTION BASED ON AN ENSEMBLE OF DEEP CONVOLUTIONAL NEURAL NETWORKS

*Balazs Harangi*

harangi.balazs@inf.unideb.hu
Faculty of Informatics, University of Debrecen, POB 400, 4002 Debrecen, Hungary

## ABSTRACT

Skin cancer is a major public health problem, with over 5 million newly diagnosed cases in the United States each year. Melanoma is the deadliest form of skin cancer, responsible for over 9,000 deaths each year. In this paper, we propose an ensemble of deep convolutional neural networks to classify dermoscopy images into three classes. To achieve the highest classification accuracy, we fuse the outputs of the softmax layers of four different neural architectures. For aggregation, we consider the individual accuracies of the networks weighted by the confidence values provided by their final softmax layers. This fusion-based approach outperformed all the individual neural networks regarding classification accuracy.

## 1. INTRODUCTION

Skin cancer is a common and locally destructive cancerous growth of the skin. It originates from the cells that line up along the membrane that separates the superficial layer of the skin from the deeper ones. As pigmented lesions occur on the surface of the skin, melanoma can be recognized early via visual inspection performed by a clinical expert. Dermoscopy is an imaging technique that eliminates the surface reflection of the skin. By removing surface reflection, more visual information can be obtained from the deeper levels of the skin.

Automatic detection of melanoma can also be addressed by using efficient automated image processing methods. As affordable mobile dermatoscopes are getting available to be attached to smart phones, the possibility for automated assessment is expected to positively influence corresponding patient care for a wide population. Given the widespread availability of high-resolution cameras, algorithms that can improve our ability to asses suspicions lesions can be of great value.

The organizers of the 2017 ISBI Challenge on Skin Lesion Analysis Towards Melanoma Detection called for participation in developing efficient methods to classify skin lesion images into three different classes. Namely, the first class contains cancerous melanoma cases, which is a malignant skin tumor derived from melanocytes. The second one labels the nevus, which is a benign skin tumor, derived from melanocytes. The last one is regards seborrheic keratosis, which is also a benign skin tumor derived from keratinocytes. According to the proper task description, the problem can be divided into two binary classification tasks as distinguishing between (a) melanoma, and (b) nevus and seborrheic keratosis; and distinguishing between (a) seborrheic keratosis, and (b) nevus and melanoma.

The challenge has provided an image set which contained a training (2,000 images), validation (150 images), and test set (600 images). The participants could apply the training (with ground-truth labels) and validation (without ground-truth labels) sets to train and fine tune their methods. The final evaluation has been performed on the test set.

## 2. PROPOSED METHODLOGY

In order to achieve the highest possible accuracy considering image classification, we have elaborated an automated method considering the ensemble of deep convolutional neural networks (DCNNs). Details on our approach are given next.

Recently, in the field of natural image classification several DCNN architectures have been published, like GoogLeNet [1], AlexNet [2], ResNet [3], VGGNet [4] beside others. Some of these DCNN architectures (GoogLeNet, ResNet) are available as pre-trained models initially trained on approximately 1 million natural images from the dataset ImageNet [5]. Thus, we can use the weights and biases from these pre-trained models. That is, if we fine-tune all the layers of these models by going on with the backpropagation using our data, then they can be applied for specific classification tasks. Other architectures as AlexNet and VGGNet contain lower number of layers, so they can start the training phase on much fewer number of images with avoiding the overfitting issue. In this way, they can be initialized so their weights and biases are not influenced by images which largely differ from skin images. The approach designed for this challenge incorporates the advances of these two type of learning solutions via composing an ensemble of the DCNNs described above.

## 2.1. Extending of the training set

The available training set contains 2,000 images with manual annotations regarding the three different classes in the following compounds: 1,372 images with nevus lesions, 254 images with seborrheic keratosis, and 374 with malignant skin tumors. The number of images in the certain classes are not sufficiently large for learning DCNNs [6]. Learning on external data sets were also allowed in this challenge, however, neither public, nor private image sets were available for us with the appropriate ground-truth. Instead, to increase the number of training images with also avoiding the overfittings of the DCNNs, we have followed the commonly proposed solution for the augmentation of the training dataset.

There are many possibilities for data augmentation, such as cropping random samples from the images or horizontally flipping or rotating them at different angles (e.g., 90°, 180°, and 270°). Using these procedures, we have generated 14,300 images from the training dataset. That is, we have increased the number of nevus images to 8,200, melanoma images to 4,600, and seborrheic keratosis to 1,500. Unfortunately, this amount is still not satisfactory for training, especially, in comparison with the dataset ImageNet [5] containing approximately 1 million natural images divided into 1,000 object categories.

## 2.2. Training and fine-tuning of the DCNNs

To create an ensemble of DCNNs, first we have to train or fine-tune them. Both actions are very time consuming procedures, so efficient implementations are also highly welcome because of the limited computational resources and the vast amount of training data. Several machine learning open source libraries are available, such as CudaConvNet [7], Torch [8], Theano [9], Caffe [10], and MatConvNet [11]. MatConvNet is a MATLAB toolbox implementing DCNNs, which are optimized for both CPU and GPU usages, as well. It provides a friendly environment for research purposes together with high computational performance thanks to the C++ and CUDA-based implementations.

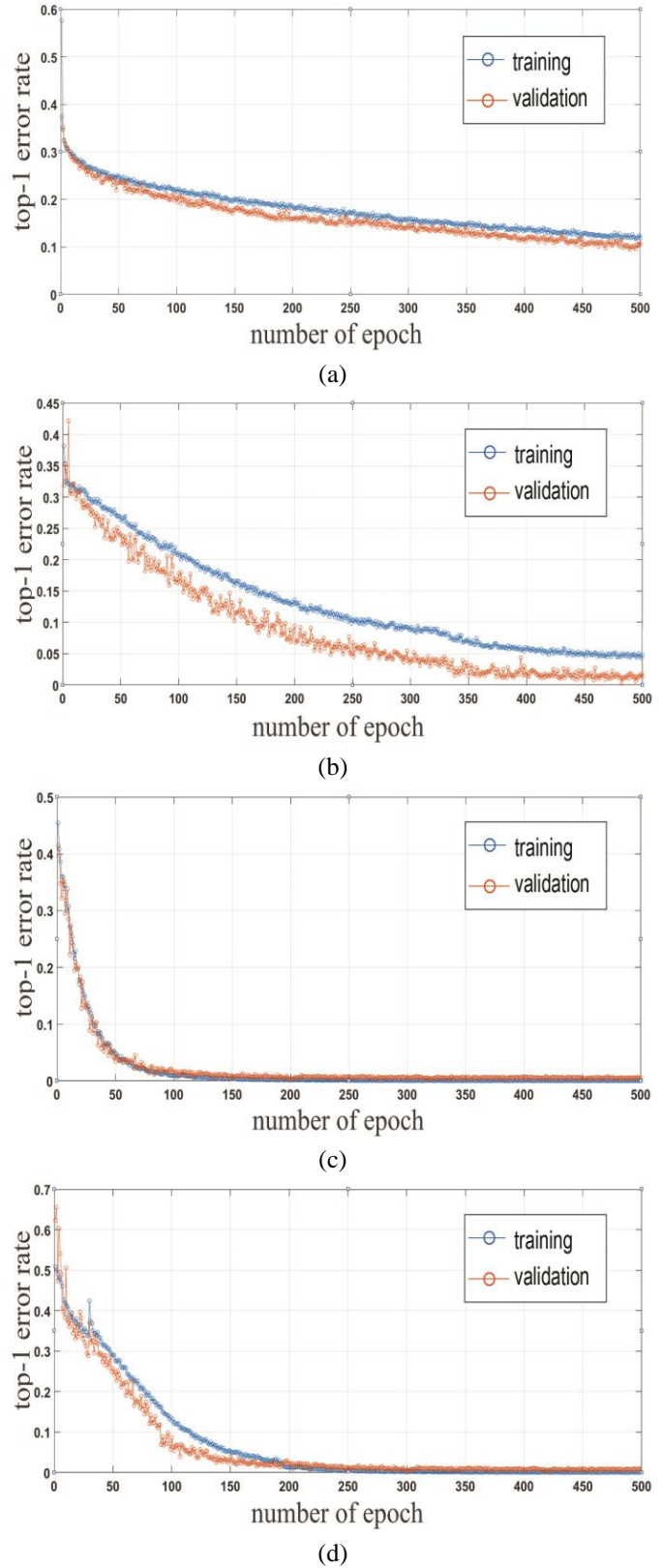

Figure 1. Training and validation results during the fine-tuning of the (a) GoogLeNet, (b) ResNet-50, (c) AlexNet and (d) VGG-VD-16 networks.

The MatConvNet toolbox provides some pre-trained DCNN models and some functions to create and initialize new neural networks. In our approach, we have considered the models GoogLeNet and ResNet-50, which were initially trained on the dataset ImageNet. Thus, we could use the weights and biases from these models trained on ImageNet, and just fine-tuned the layers of them. Moreover, we have created the AlexNet and VGG-VD-16 neural network architectures, which had weights and biases initialized by random values.

The fine-tuning and training has been performed on a computer equipped with an NVIDIA TITAN X GPU card with 7 TFlops of single precision, 336.5 GB/s of memory bandwidth, 3,072 CUDA cores, and 12 GB memory. The convolutional filters of the DCNNs were found by a stochastic gradient descent algorithm iterated through 500 training epochs per neural network.

During these fine-tuning and training processes, the results of backpropagation and the classification of the validation images after each epoch was determined (see Figure 1) using the top-1 error rate [2]. It can be observed that the overfitting problem is successfully resolved, since the validation curves do not increase after overfitting the CNN model.

**2.2. Ensemble of the DCNNs**

After training the DCNNs on the augmented dataset, we have composed an ensemble from them to increase the overall accuracy of classification. The basic fusion model of the networks has been selected to be the simple majority voting one, when the final class label of an input image is derived as the majority of the ones provided by the individual DCNNs. As a simple demonstrative example, imagine that we have an ensemble of three classifiers: $\{h_1, h_2, h_3\}$ and consider a new case $x$ to be classified. If the three classifiers are uncorrelated, then when $h_1(x)$ is wrong, $h_2(x)$ and $h_3(x)$ may be correct, so majority voting will correctly classifies $x$ [12].

In our decision, we must handle the problem, when two classes receives the same number of votes, that is, we have a tie in majority voting. To resolve this case and to boost classification accuracy, we have assigned weights to the voters and considered a weighted majority voting ensemble of DCNNs. To adjust the proper weights, we have determined the individual accuracies of the GoogLeNet, AlexNet, ResNet-50, and VGG-VD-16 based on the averages of their area under the receiver operating characteristic curves (AUCs) considering the melanoma and the seborrheic keratosis detection results on the validation set (150 images). By doing so, we have obtained the weights $w_{GLN} = 0.895$, $w_{AN} = 0.851$, $w_{RN} = 0.846$, and $w_{VGG} = 0.862$ for the GoogLeNet, AlexNet, ResNet-50, and VGG-VD-16, respectively. We can also interpret these weights as information about the reliability of the corresponding DCNNs.

The softmax layer of a DCNN determines a confidence value for each class, and in a simple case, a neural network assigns its final label to the input image based on selecting the class with maximal confidence value. Regarding our approach, these confidence values are also to be taken into consideration in our majority voting scheme. Namely, if $p_{i,j}$ denotes the confidence value assigned by the $i^{th}$ classifier to the $j^{th}$ class with $i \in \{$GoogLeNet, AlexNet, ResNet-50, VGG-VD-16$\}$, and $j \in \{$melanoma, nevus, seborrheic keratosis$\}$, then $Pmax_i = \max_j(p_{i,j})$ gives the highest confidence value provided by the $i^{th}$ classifier with the corresponding class label $C_i$. Then, the final class label $\mathcal{FC}$ can be derived using weighted majority voting as

$$\mathcal{FC} = \max_j \sum_{i|C_i=j} w_i Pmax_i.$$

That is, we consider that class label as final, which gets the largest weighted sum. Notice that, having a tie has a very low probability using this scheme.

**3. EXPERIMENTAL RESULTS**

According to the official evaluation of our proposed ensemble-based system, it has achieved an overall score 0.932 on the validation set, which contained 150 images. This overall score has been calculated as the average of the AUC corresponding to the melanoma and the seborrheic keratosis classification results. Regarding the AUC, the sensitivity and specificity have been calculated at the confidence threshold 0.5. Moreover, specificity has been evaluated at the sensitivity levels 82%, 89%, and 95%. The quantitative results are summarized in Table I, where M/SK denote melanoma/seborrheic keratosis, while ACC/AP/SE/SP stand for accuracy/average precision/sensitivity/specificity, respectively.

TABLE I. EXPERIMANTAL RESULTS ON THE VALIDATION SET.

| AVG_ACC | M_ACC | SK_ACC |
|---------|-------|--------|
| 0.893 | 0.867 | 0.920 |
| AVG_AUC | M_AUC | SK_AUC |
| 0.932 | 0.899 | 0.964 |
| AVG_AP | M_AP | SK_AP |
| 0.834 | 0.753 | 0.915 |
| AVG_SE | M_SE | SK_SE |
| 0.600 | 0.367 | 0.833 |
| AVG_SP82 | M_SP82 | SK_SP82 |
| 0.894 | 0.833 | 0.954 |
| AVG_SP89 | M_SP89 | SK_SP89 |
| 0.850 | 0.775 | 0.926 |
| AVG_SP95 | M_SP95 | SK_SP95 |
| 0.659 | 0.475 | 0.843 |
| AVG_SP | M_SP | SK_SP |
| 0.973 | 0.992 | 0.954 |

## 4. CONCLUSION

In this paper, we have proposed an ensemble of deep convolutional neural networks (DCNNs), which has outperformed the accuracy of the individual DCNN components in a classification task regarding skin lesions. To derive the final class label of the input image, we have considered the individual accuracies of the DCNNs and incorporated these figures in a weighted majority voting scheme as a decision rule.

## ACKNOWLEDGMENT

This work was supported in part by the projects GINOP-2.1.1-15-2015-00376 and VKSZ 14-1-2015-0072, SCOPIA: Development of diagnostic tools based on endoscope technology supported by the European Union, co-financed by the European Social Fund. We are also grateful to Laszlo Kovacs for making a graphic card (provided by NVIDIA for his micro HPC development activities) available for our experiments.